\documentclass[runningheads]{llncs}
\usepackage[T1]{fontenc}
\usepackage{graphicx}
\usepackage[misc]{ifsym}
\usepackage{fontawesome5}
\usepackage{subcaption}
\usepackage{calc}
\usepackage{amssymb}
\usepackage{amstext}
\usepackage{amsmath}
\usepackage{booktabs}
\usepackage{color}
\begin{document}
\title{From Predictions to Explanations: Explainable AI for Autism Diagnosis and Identification of Critical Brain Regions}
%From Prediction to Explanation: Discovering ASD-Related Brain Regions through XAI Techniques
%Interpretable Deep Learning for Autism Diagnosis: Identifying Key Brain Regions with Explainable AI
\titlerunning{Explainable AI for Autism Diagnosis}
\author{Kush Gupta\inst{1}\orcidID{0009-0008-9930-6435} \and
Amir Aly\inst{1}\orcidID{0000-0001-5169-0679} \and
Emmanuel Ifeachor\inst{1}\orcidID{0000-0001-8362-6292} \and
Rohit Shankar \inst{1}\orcidID{0000-0002-1183-6933}}

\authorrunning{Kush Gupta et al.}
\institute{University of Plymouth, Plymouth, UK \\
\email{kush.gupta@plymouth.ac.uk \textsuperscript{\faEnvelope[regular]}, amir.aly@plymouth.ac.uk,}\\
\email{E.ifeachor@plymouth.ac.uk, rohit.shankar@plymouth.ac.uk}\\
}

\maketitle              % typeset the header of the contribution
\begin{abstract}
Autism spectrum disorder (ASD) is a neurodevelopmental condition characterized by atypical brain maturation. However, the adaptation of transfer learning paradigms in machine learning for ASD research remains notably limited. In this study, we propose a computer-aided diagnostic framework with two modules. This chapter presents a two-module framework combining deep learning and explainable AI for ASD diagnosis. The first module leverages a deep learning model fine-tuned through cross-domain transfer learning for ASD classification. The second module focuses on interpreting the model’s decisions and identifying critical brain regions. To achieve this, we employed three explainable AI (XAI) techniques: saliency mapping, Gradient-weighted Class Activation Mapping, and SHapley Additive exPlanations (SHAP) analysis. This framework demonstrates that cross-domain transfer learning can effectively address data scarcity in ASD research. In addition, by applying three established explainability techniques, the approach reveals how the model makes diagnostic decisions and identifies brain regions most associated with ASD. These findings were compared against established neurobiological evidence, highlighting strong alignment and reinforcing the clinical relevance of the proposed approach.

\end{abstract}
\section{Introduction}

Autism spectrum disorder (ASD) is a neurodevelopmental condition characterized by atypical brain maturation \cite{nur}. Core manifestations involve persistent deficits in social communication and interaction, alongside restricted patterns of interest and repetitive behaviours \cite{maenner2023prevalence}. Furthermore, individuals diagnosed with ASD frequently present with co-occurring traits, including delays in both linguistic and motor skill acquisition, heightened levels of anxiety and stress, and atypical emotional or mood responses. ASD diagnoses have surged dramatically in recent decades, rising from ~1\% to nearly 3\% of the population, reflecting a staggering 787\% increase over twenty years \cite{saito2020prevalence}. Current estimates indicate that approximately 1 in 35 children in the United States (US) receive an ASD diagnosis, with males demonstrating a markedly higher susceptibility; the male-to-female ratio approaches 3:1 \cite{shaw2025prevalence}.  In the UK alone, over 200,000 individuals now endure lengthy waiting lists for evaluation \cite{russell2022time}. This exponential growth, fuelled significantly by rising adult diagnoses alongside resource-intensive assessment protocols, has precipitated a diagnostic crisis. Establishing an ASD diagnosis presents considerable challenges due to the absence of distinctive physical markers. Consequently, clinicians predominantly rely on standardized diagnostic instruments, such as the Autism Diagnostic Observation Schedule, Second Edition (ADOS-2), the criteria outlined in the Diagnostic and Statistical Manual of Mental Disorders, Fifth Edition (DSM-5), and the International Classification of Diseases, 11th Revision (ICD-11) to evaluate diagnostic probability \cite{Sharda2016,JENNINGSDUNLAP2019496}. 

These traditional methods necessitate extensive clinical expertise and involve protracted observational periods, often requiring 4-6 hours per assessment \cite{WANKHEDE2024104241}. This contributes to substantial delays, with diagnoses typically occurring between ages 4-6 years in the United States, considerably later than the optimal intervention window of 2-3 years. Furthermore, these methods exhibit inherent subjectivity, as diagnostic accuracy remains heavily dependent on clinician experience and training, resulting in diagnostic agreement rates as low as 70\% \cite{Thakkar}. Resource constraints exacerbate these issues, particularly in low-income regions where mental health specialist availability may be as limited as 1 per 100,000 individuals \cite{nilsen2022accelerating}. Additionally, cultural and gender biases persist within traditional frameworks, leading to under-diagnosis in female, Hispanic, and Black populations \cite{Thakkar}. Consequently, these conventional diagnostic methodologies for ASD have significant limitations that impede reliable screening and effective intervention.

Furthermore, substantial clinical consequences are observed when ASD diagnoses are delayed beyond the optimal intervention window of 2-3 years. Late-diagnosed children exhibit significantly worsening trajectories of emotional, behavioural, and social difficulties (EBSDs) throughout adolescence compared to those diagnosed earlier. By age 14, these individuals demonstrate markedly higher levels of internalising problems, conduct issues, hyperactivity, and peer relationship challenges, even after controlling for factors such as IQ, gender, and maternal education \cite{mandy2022mental}. Additionally, diagnostic delays contribute to increased psychiatric co-morbidities, as prolonged unmet support needs exacerbate anxiety, depression, and self-injurious behaviours before diagnosis \cite{vu2023increased}. Crucially, late diagnosis prevents access to early intensive intervention during critical neurodevelopmental periods, resulting in reduced treatment efficacy and poorer long-term outcomes in communication, adaptive functioning, and independence \cite{thomas2022symptoms}.

Moreover, these diagnostic delays create significant economic and systemic burdens. Analysis of commercially insured children reveals that those experiencing longer time-to-diagnosis (TTD) incur approximately double the healthcare costs in the year preceding diagnosis compared to those with shorter TTD (\$5,268 vs \$2,525 for younger cohorts). This is primarily driven by a 1.5 to 2-fold increase in healthcare visits as families navigate protracted diagnostic pathways \cite{vu2023increased}. Simultaneously, delayed diagnoses strain educational systems and specialist services, as undiagnosed children often require crisis-driven support rather than preventative interventions. Societally, late diagnosis perpetuates health inequalities, with diagnostic disparities particularly affecting females, ethnic minorities, and children from socio-economically disadvantaged backgrounds due to resource limitations and cultural biases inherent in traditional assessment approaches \cite{Leslie2024magazine}. Traditional diagnostic limitations necessitate prolonged specialist-dependent evaluations (e.g., 4-6 hours for assessments) and demonstrate concerning subjectivity, with inter-clinician agreement rates as low as 70\% \cite{song2019use}. These approaches frequently miss subtle early indicators, particularly in children with co-occurring conditions like ADHD or in those with higher masking capabilities \cite{mandy2022mental}. 

Artificial intelligence (AI) methodologies are addressing these systemic shortcomings through multifaceted innovations in ASD diagnosis.  Machine learning algorithms are applied to existing diagnostic instruments to identify predictive item subsets, drastically reducing assessment times without compromising accuracy \cite{Leslie2024magazine}. Natural language processing (NLP) enables automated analysis of vocal patterns and social communication features, reducing observational subjectivity. AI-powered tools analyse subtle behavioural signatures not captured by conventional methods. Tablet-based applications assessing motor kinematics during drawing tasks differentiate ASD from typical development, providing quantifiable motor biomarkers \cite{Leslie2024magazine}. 
Computer vision algorithms extract micro-behavioural features (e.g., eye contact frequency, facial expressivity) from brief home videos, enabling remote assessment. 

Research in this domain increasingly prioritizes quantifiable neuroimaging techniques, particularly functional Magnetic Resonance Imaging (fMRI), recognized as a prominent modality for ASD identification \cite{Klin2018-xw}. AI-driven analysis of neuroimaging data facilitates the identification of physiological indicators long before behavioural symptoms manifest conclusively. Deep learning models detect microstructural white matter alterations in diffusion tensor imaging (DTI) and functional connectivity patterns in resting-state fMRI, achieving good classification accuracies in children under 24 months \cite{wankhede2024leveraging}.

%fMRI offers access to objective neurophysiological biomarkers \cite{traut2022insights}, thereby diminishing dependence on subjective clinical interpretation.

Over the past two decades, computer-assisted diagnosis (CAD) systems leveraging AI have demonstrated significant scientific and clinical utility. Neural architectures are frequently employed to derive condensed, fixed-dimensional feature embeddings from extensive public datasets. These representations are subsequently adapted via knowledge transfer methodologies to refine models for diverse research applications, enhancing cross-domain generalization capabilities. Emerging evidence positions neural networks and transfer learning as viable instruments for mental illness prevention strategies \cite{Durstewitz2019-ow}. Nevertheless, the adaptation of transfer learning paradigms to autism spectrum disorder (ASD) research remains notably limited. This paucity arises partly from ASD's heterogeneous neurodevelopmental nature, marked by intricate cognitive phenotypes \cite{cao2023commentary}. Consequently, substantial obstacles persist in acquiring comprehensive ASD datasets and establishing robust CAD frameworks. The Autism Brain Imaging Data Exchange (ABIDE) consortium \cite{abidei} aggregated functional Magnetic Resonance Imaging (fMRI) data encompassing 539 ASD individuals and 573 neurotypical controls. The present study utilizes the ABIDE dataset complemented by fMRI data from the Child Mind Institute's Healthy Brain Network (CMI-HBN) initiative \cite{cmi_hbn}. 

%Furthermore, while current research utilizing AI-based methods demonstrates ASD classification capabilities, a critical limitation persists: these studies frequently lack transparent mechanistic interpretations of their outcomes. Result interpretability holds substantial clinical relevance, as it deepens practitioners' understanding of algorithmic decision pathways and augments diagnostic reasoning. Explainable Artificial Intelligence (XAI) addresses the fundamental disconnect between AI's operational opacity and is imperative for human-interpretable insights. 

AI methods, particularly complex deep learning models, frequently function as "black boxes," where decision-making processes remain opaque. This opacity presents substantial barriers in high-stakes domains such as healthcare, where understanding the rationale behind diagnostic or therapeutic recommendations is clinically imperative. When AI systems provide outputs without transparent reasoning, their utility is diminished, as healthcare practitioners cannot independently verify the validity or pathological basis of conclusions. In ASD diagnosis, for instance, traditional machine learning models may achieve high classification accuracy yet fail to elucidate which behavioural or neuroanatomical features drove specific assessments, creating a fundamental disconnect between AI’s operational mechanisms and clinicians’ need for interpretable insights. Explainable Artificial Intelligence (XAI) directly addresses this limitation by rendering algorithmic processes auditable and comprehensible, thereby transforming AI from an inscrutable tool into a collaborative partner in clinical reasoning. In ASD diagnostics, where early intervention critically influences developmental trajectories, opaque models risk rejection by practitioners despite technical accuracy. XAI methodologies, such as Local Interpretable Model-agnostic Explanations (LIME) and Shapley Additive Explanations (SHAP), demystify AI outputs by identifying decisive input features—for example, highlighting specific facial metrics in image-based ASD screening or quantifying the influence of genetic markers on risk predictions \cite{atlam2025automated}. Result interpretability holds substantial clinical relevance, as it deepens practitioners' understanding of algorithmic decision pathways and augments diagnostic reasoning.

Beyond trust, XAI actively enhances diagnostic accuracy and therapeutic personalisation. By elucidating feature contributions, clinicians can prioritise high-impact variables during assessments—such as specific items in the Autism Diagnostic Observation Schedule (ADOS-2) assessments. Additionally, XAI supports personalised intervention strategies by clarifying how patient-specific factors (e.g., genetic variants or neuroimaging abnormalities) modulate risk predictions or treatment responses. This capability transforms AI from a static classifier into a dynamic tool for precision medicine, where explanations inform not only diagnoses but also individualised management plans. This transparency fosters confidence in AI-assisted diagnoses and facilitates smoother integration into existing clinical workflows.

Our primary research objective was the development of a deep learning (DL) model capable of achieving accurate ASD diagnosis while ensuring the provision of interpretability and transparent decision pathways in its outputs. Through the integration of explainable methodologies with our DL module, insights regarding contributory diagnostic mechanisms can be derived by medical practitioners and investigators. Additionally, significant brain regions could be determined by the various XAI methods. The two modules of our framework are summarized below:

% Our primary objective was to design a DL model capable of accurately identifying ASD while simultaneously offering interpretability and transparency in its outcomes. Incorporating explainable techniques into DL frameworks allows clinicians and researchers to gain insight from the underlying mechanisms contributing to the ASD diagnosis, as well as to determine important brain regions identified by the various XAI methods. 

\begin{enumerate}
    \item Recognizing the challenge of training deep neural networks without extensive fMRI datasets, we implemented inter-domain transfer learning combined with knowledge distillation (KD) loss.   
    %This approach repurposes models initially trained for one task as foundations for related objectives—significantly reducing data requirements \cite{survey_tl}. 
    The first module of our framework \cite{gupta2025cross} leverages TinyViT \cite{wu2022tinyvit}, a novel family of compact vision transformers evolved from the original ViT architecture \cite{vit}.
    %These models effectively overcome limitations in both CNNs and traditional ViTs by processing images as patch sequences rather than local regions. Crucially, their window-attention mechanism evaluates all patches concurrently, enabling superior capture of global context and long-range dependencies, proving essential for tasks requiring holistic understanding. Unlike CNNs, TinyViTs exhibit minimal architectural bias toward local spatial patterns. This flexibility allows learning more abstract, complex data representations. While standard ViTs demand substantial computational resources, TinyViT addresses this via efficient pre-training on ImageNet \cite{deng2009ImageNet} using rapid knowledge distillation. Here, expertise from larger models transfers to compact counterparts, granting smaller architectures the benefits of massive pretraining datasets \cite{yosinski2014transferable}. 
    %We initiate with TinyViT’s ImageNet-pretrained foundation, where distilled knowledge during pretraining establishes vital feature representations. The architecture automatically scales down large pre-trained models while respecting hardware constraints, boosting performance without demanding excessive labeled data.
    %This efficiency accelerates deployment and democratizes access for resource-limited settings. Finally, 
    We fine-tune the model on our specialized fMRI domain. This critical step preserves valuable, pre-trained knowledge while adapting to domain-specific patterns—an essential strategy in healthcare, where large-scale data sharing remains challenging. %Clinicians gain not just a tool, but an adaptable solution that evolves with their diagnostic needs.
    
    \item The second module in our framework is XAI methods.  We employed three different XAI methods, namely Saliency maps \cite{saliency}, Grad-CAM \cite{gradcam}, and SHAP \cite{shap}, to identify critical brain regions when diagnosing ASD. We have explicitly utilized XAI methodologies to enhance model transparency and comprehensibility. By making AI predictions explainable, clinicians can comprehend the rationale underlying automated decisions, fostering clinically meaningful analysis and establishing essential trust. %\cite{haq2020intelligent}.
    With the XAI in our framework, we were able to identify and highlight important brain regions
    critical for ASD diagnosis. Furthermore, the brain areas identified by our approach corroborate with the recent neurobiological findings \cite{robertson2014global,xiao2023atypical,monk2009abnormalities,vidya2025}.
    %such as the calcarine sulcus, which corresponds to Brodmann Area (BA)\cite{broadmann} (17), a center for visual processing. The Insula, BA (13), which contributes to cognitive processes such as decision-making, working memory, and attention. Also, other areas that were common between the saliency maps and SHAP are the parietal lobe, which corresponds to BA (5), which is responsible for somatosensory and visual systems to guide movement and spatial awareness. Emerging evidence has highlighted the involvement of the primary sensory cortex; notably, genomic analyses reveal that BA (17)- the primary visual cortex—exhibits the most pronounced anomalies in individuals with autism spectrum disorder \cite{gandal2022broad,vidya2025}. 
\end{enumerate}

%The structure of this study has been organized as follows: a review of relevant literature is presented in the subsequent section. The datasets employed in the investigation are described in section (III). A comprehensive overview of the proposed framework is provided in section (IV), followed by a description of the experimental setup and implementation specifics in section (V). The results derived from the study are reported in section (VI), and key insights are discussed in Section (VII). Finally, the conclusions formulated from this research are outlined in section (VIII).

This chapter provides an overview of our framework, which comprises cross-domain transfer learning and XAI methodologies for ASD diagnosis. It discusses the datasets used and the methodological approach. Further, it outlines the experimentation settings and implementation details. Finally, it discusses the obtained results and our key findings. % and explores future research directions.

\section{Related Work}
Current clinical ASD assessments remain heavily reliant on behavioural observation and patient history, approaches with inherent diagnostic constraints. These methods detect atypical social communication patterns that frequently become apparent only after the condition is entrenched \cite{mccarty2020early}. AI-based approaches are being increasingly developed to address these shortcomings. Machine learning and deep learning algorithms are capable of analysing vast and complex datasets, including behavioural video data, speech patterns, neuroimaging, and genetic profiles. These models can detect patterns and features that may not be apparent to human observers, thus enhancing the objectivity of diagnostic outcomes \cite{duda2016use,gupta2025multi}. 

Recent research has revealed that fMRI offers access to objective neurophysiological biomarkers \cite{traut2022insights}, thereby diminishing dependence on subjective clinical interpretation.
Specifically, the resting-state fMRI (rs-fMRI) now offers transformative potential for ASD diagnostics. The field’s growing interest stems partly from the Autism Brain Imaging Data Exchange (ABIDE) \cite{abidei}, which pooled functional and structural neuroimaging data across 17 international sites, creating unprecedented research opportunities. This collaborative resource empowers us to reimagine how we detect and understand autism. Over the past decade, the ABIDE dataset has served as the cornerstone for numerous ASD studies \cite{heinsfeld2018identification,iidaka2015resting}. Researchers often focus on specific demographic subgroups within ABIDE, allowing us to see how autism manifests differently across populations, revealing nuances that broader analyses might miss. For instance, \cite{iidaka2015resting} proposed a probabilistic neural network approach using rs-fMRI scans from 312 young ASD individuals and 328 neurotypical controls (all under age 20), reporting 90\% classification accuracy. Meanwhile, the study \cite{plitt2015functional} examined two targeted cohorts: 118 males (59 ASD/59 TD) and 178 individuals age-matched and IQ-matched (89 ASD/89 TD). Their model achieved 76.67\% accuracy, demonstrating that subgroup analysis can yield imperative insights despite smaller sample sizes.

To improve the ASD diagnosis, researchers worldwide have started harnessing the power of neural architectures like Deep Neural Networks (DNNs), Long Short-Term Memory (LSTM) networks, and Auto-encoders to decode ASD’s neural signatures. Consider Brown et al. \cite{brown2018connectome}, who designed an element-wise DNN layer incorporating structural priors. Their model classified 1013 subjects (539 controls / 474 ASD) at 68.7\% accuracy—a promising step toward translating scans into clinical insights. Yet these approaches share a constraint: reliance on hand-engineered feature extractors that struggle to generalize across new patients. With a sudden upsurge in the incidence of ASD cases, the variability across the data is also increasing. Since these methods rely on hand-engineered features, they would struggle to perform and generalize across the new data. %This point raises a critical reminder that scalability remains as crucial as accuracy in real-world diagnostics.

Convolutional Neural Networks (CNNs) have been mainly utilized within CAD frameworks, leveraging fMRI data from the ABIDE repository to distinguish autistic individuals from typically developing controls (TC) \cite{husna2021functional}. \cite{manaswi2018deep,sherkatghanad2020automated} studies employed CNN architectures to extract discriminative features for ASD/TC classification. Other teams achieved similar milestones: \cite{sherkatghanad2020automated} reached 70.22\% accuracy with CNNs on ABIDE data, whereas \cite{dvornek2018combining} also reported ~70\% accuracy using similar architectures. As foundational frameworks in deep learning, CNNs excel particularly in visual pattern recognition. Notwithstanding their prevalence, CNNs exhibit inherent constraints: their convolutional layers operate via localized receptive fields, prioritizing regional pixel relationships. While effective for capturing spatial hierarchies, this design inherently restricts the modelling of long-range dependencies or global contextual information. Additionally, CNNs possess a pronounced architectural inductive bias favouring translational invariance and locality assumptions. Such bias impedes the learning of highly abstract or non-local feature representations. 

To overcome these constraints, researchers started to utilize the transformer-based architectures such as the ViT architecture \cite{vit}. Transformers are increasingly favoured over CNNs because of their enhanced global contextual modelling capabilities. The input data is processed as sequential patch arrays within transformer-based architectures, with long-range dependencies between spatially separated patches being captured through self-attention mechanisms. This approach facilitates the assimilation of comprehensive contextual information across complete input data, overcoming the inherent locality constraints of CNNs where fixed receptive fields and inductive spatial biases are relied upon. Consequently, more intricate and abstract data representations are learned, as architectural presuppositions concerning spatial relationships are not imposed. Inter-patch relationships are dynamically weighted by attention mechanisms, enabling complex interactions to be modelled irrespective of positional proximity. One of the limitations arises from their data-intensive nature, necessitating extensive image datasets for robust training. Training transformer-based models, de novo, incurs substantial computational overhead, extended training durations, and dependency on specialized hardware infrastructure. However, this limitation could be easily managed by utilising the cross-domain transfer learning paradigm.
This approach repurposes models initially trained for one task as foundations for related objectives—significantly reducing data requirements \cite{survey_tl}. The models pre-trained on large datasets such as ImageNet \cite{deng2009ImageNet} could be used as base models for fine-tuning on the domain-specific datasets. The application of cross-domain transfer learning aids in the transfer of knowledge from comprehensive natural image datasets to the specialized field of brain imaging, thereby enabling the deployment of transformer-based models even in areas where data is scarce.

%Clinicians and researchers face a persistent hurdle in autism diagnosis: pinpointing meaningful biomarkers within intricate fMRI data. 

A critical challenge in current ASD deep learning research involves the "black box" nature of diagnostic models, which fail to reveal the neuroanatomical basis for their classifications \cite{salahuddin2022transparency}. Compounding this opacity, studies employing interpretability techniques typically utilize single methods without comparative analysis. More concerningly, few validate their findings against established neuroscientific knowledge, undermining both reliability and clinical translation potential. In life-critical domains like medical diagnostics, model transparency is paramount in understanding the rationale behind algorithmic decisions to establish essential trust towards clinical outcomes. XAI empowers researchers to not only identify disorders under specific conditions but to decipher the causal pathways driving these predictions. These interpretability methods transform data into actionable clinical intelligence, enabling practitioners to deliver precisely calibrated interventions grounded in mechanistic understanding. The integration of XAI with biomedical analytics further catalyses precision medicine initiatives. By elucidating how individual genetic variations influence disorder manifestation, these approaches enhance diagnostic specificity while unlocking personalized therapeutic pathways. Given the profound heterogeneity of neurodevelopmental conditions, such patient-tailored frameworks could transform diagnostic and management paradigms across healthcare systems.

While explainable AI (XAI) shows promise in medical imaging—powering cervical cancer screening through gradient-based methods (Grad-CAM, Layer-wise Relevance Propagation) \cite{shyamalee2024automated}, enhancing melanoma detection with Grad-CAM variants \cite{gamage2024melanoma}, and advancing glaucoma diagnosis via visualization techniques —these successes remain concentrated in domains where clinically relevant features are visually discernible. The ASD diagnosis domain still faces the fundamental challenge of making fMRI-driven ASD diagnostics both interpretable and neurologically grounded.

% talk about a few papers that deals with XAI in ASD.
These limitations underscore the urgent need for an explainable AI-based CAD system that enables more efficient and accurate identification of ASD. Furthermore, the CAD system should be able to provide additional insights into the diagnosis, allowing clinicians to make an informed and prompt decision to plan a more effective intervention early. 

Despite these advances, existing studies rarely integrate cross-domain transfer learning with multi-method XAI approaches validated against neuroscientific evidence. This chapter seeks to address the existing gap by presenting a novel framework. The framework comprises two main modules: the integration of cross-domain transfer learning aimed at improving diagnostic precision, and the utilization of various XAI techniques to elucidate interpretability in ASD neuroimaging. This framework offers both a consensus and new perspectives on the neuropathology associated with ASD.

\section{Datasets}
Functional Magnetic Resonance Imaging (fMRI) represents a cornerstone neuroimaging technique that captures dynamic brain activity through haemodynamic changes \cite{lindquist2008statistical}. This methodology partitions the brain into volumetric pixels (voxels), each generating a temporal signature reflecting neural activation patterns. Our investigation specifically leverages resting-state fMRI (rs-fMRI), where subjects maintain passive alertness without performing structured tasks—either fixating on a crosshair or keeping eyes closed while permitting spontaneous cognition \cite{Gonzalez-Castillo2021-es}. This protocol eliminates motor/perceptual demands, making it particularly valuable for studying neurodevelopmental conditions. Our analysis utilizes rs-fMRI data from both the ABIDE \cite{abidei} repository and the CMI-HBN \cite{cmi_hbn} initiative.

The resting-state fMRI (rs-fMRI) data from the Autism Brain Imaging Data Exchange (ABIDE) repository, which permits unrestricted academic use. This curated dataset comprises 1112 rs-fMRI scans acquired across 17 international sites, including 505 autistic individuals and 530 typical controls. ABIDE equips mean time-series data derived from seven distinct brain atlases. %Following established preprocessing protocols via the C-PAC pipeline \cite{craddock2013neuro}, we implemented rigorous motion artifact exclusion: empirical evidence indicates Framewise Displacement (FD) values > 0.2mm significantly compromise fMRI data integrity \cite{Power2014-uv,Power2012-eh}. 
%Consequently, scans exceeding mean FD >0.2mm were excluded, yielding final cohorts of 424 ASD patients and 510 healthy controls. %Site-specific demographic distributions are detailed in Figure \ref{fig:abide_i_table}.

% \begin{figure}[!h]
%   \centering
%    %{\epsfig{file = Figures/, width = 5.5cm}}
%     \includegraphics[width=\linewidth]{Figures/chart.png}
%   \caption{Site-wise distribution of retained participants per diagnostic cohort following framewise displacement (FD) filtration in ABIDE. \cite{gupta2025cross}} \label{fig:abide_i_table} \centering.
% \end{figure}

The Healthy Brain Network (HBN) initiative addresses critical gaps in developmental neuroscience by establishing a large-scale, pan-diagnostic repository capturing the heterogeneity of mental health and learning profiles. Spearheaded by the Child Mind Institute, this restricted-access biobank aggregates multimodal data from 10,000 New York City participants (ages 5-21), encompassing psychiatric assessments, behavioural/cognitive metrics, lifestyle factors (diet, fitness), multimodal neuroimaging (including MRI/EEG), audiovisual recordings, genetic data, and actigraphy. For this investigation, we utilized rs-fMRI scans from 359 ASD and 359 neurotypical subjects.% across four HBN collection sites: Staten Island (SI), Rutgers University Brain Imaging Center (RUBIC), Citigroup Cornell Brain Imaging Center (CBIC), and City College of New York (CUNY).

%Consistent preprocessing pipelines were applied to both ABIDE and HBN datasets, including: slice timing correction, motion realignment, nuisance signal regression, low-frequency drift removal, and intensity normalization. Crucially, each imaging site employed divergent acquisition protocols—varying in repetition time (TR), echo time (TE), voxel dimensions, scan duration (volume count), and resting-state conditions (eyes open/closed)—necessitating rigorous harmonization procedures.

\begin{figure*}[!h]
  \centering
   %{\epsfig{file = Figures/, width = 5.5cm}}
    \includegraphics[width=0.9\linewidth]{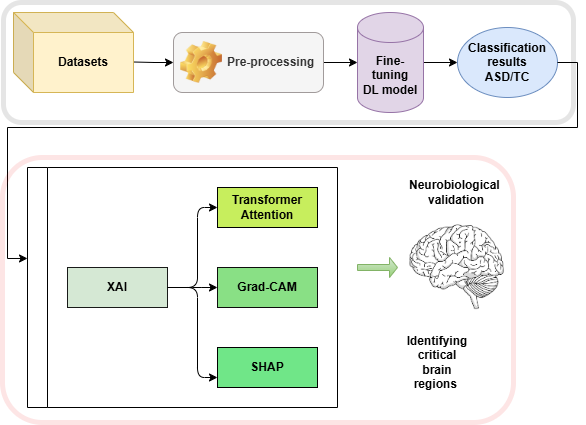}
  \caption{Schematic representation of the proposed dual-module framework. The top segment (black outline) depicts the ASD classifier module, conceptualized as an opaque deep learning architecture. The lower segment (red outline) highlights the integrated explainable AI (XAI) module, which provides insights into critical brain regions for ASD.} \label{fig:framework} \centering.
\end{figure*}

\section{Methodology}
% This section details our dual-module diagnostic framework as illustrated in Figure (\ref{fig:framework}). The ASD classifier module (top, black outline) employs a computationally efficient TinyViT architecture that achieves vision transformer-level performance with minimal parameters despite limited neuroimaging data. This operationally opaque "black box" module delivers robust ASD detection capabilities. Architectural specifics appear in Figure (\ref{fig:tinvit_arch}), featuring a lightweight MLP decoder. The explainable AI (XAI) module (bottom, red outline) integrates three complementary interpretability methods to identify ASD-relevant neuroanatomical brain regions. Our analysis revealed consensus regions across these techniques, highlighting critical brain regions for ASD diagnosis. Subsequent subsections elaborate on each framework module.

\begin{figure*}[!h]
  \centering
   %{\epsfig{file = Figures/, width = 5.5cm}}
    \includegraphics[width=\linewidth]{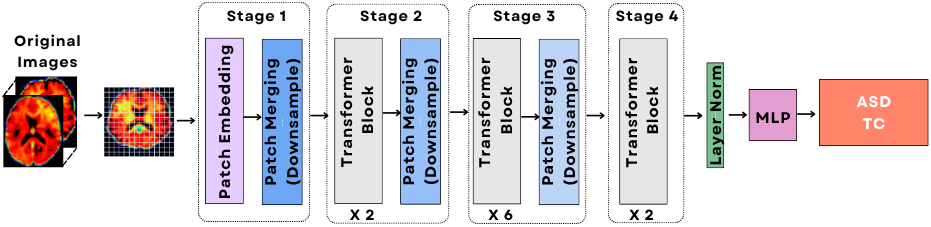}
  \caption{Architectural schematic of TinyViT modules deployed for autism spectrum disorder (ASD) classification. \cite{gupta2025cross}} \label{fig:tinvit_arch}
\end{figure*}

Our diagnostic framework consists of two modules as illustrated in Figure (\ref{fig:framework}). The first module (top, black outline) is a deep-learning classifier that employs a computationally efficient TinyViT architecture that achieves vision transformer-level performance with minimal parameters despite limited neuroimaging data. This operationally opaque "black box" module delivers robust ASD detection capabilities. Architectural specifics appear in Figure (\ref{fig:tinvit_arch}). The second module (bottom, red outline) consists of three explainable AI (XAI) methods to interpret and identify ASD-relevant neuroanatomical brain regions. Subsequent subsections elaborate on each framework module.

\subsection{First module} \label{subsec:firstcom}
% The first module of our framework is a Deep Learning (DL) model \cite{gupta2025cross}, based on the current state-of-the-art transformer architecture. More specifically, we fine-tuned the TinyViT model as an ASD classifier. Structurally, TinyViT is adapted from the hierarchical vision transformer architecture. Particularly, four sequential stages constitute the base model. Image dimensionality is progressively reduced across stages, mirroring the Swin transformer approach \cite{liu2021swin}. Patch embedding is constructed through dual convolutional layers employing a kernel size of 3 and a stride of 2. Stage 1 incorporates lightweight MBConvs \cite{howard2019searching} and down-sampling blocks, where high inductive biases enable effective low-level feature extraction. The subsequent three stages are composed of transformer blocks with window attention to optimize computational efficiency. Localized information integration is facilitated by attention biases \cite{graham2021levit} and a $3\times3$ depth-wise convolution inserted between the MLP and attention mechanisms \cite{codella2019skin}. Residual connections \cite{he2016deep} are implemented throughout Stage 1 blocks, including MLP and attention components. Activation functions universally employ GELU non-linearity \cite{hendrycks2016gaussian}. Normalization operations are differentiated: convolutional layers utilize BatchNorm \cite{ioffe2015batch}, while linear operations apply LayerNorm \cite{lei2016layer}.

The first module of our framework is a Deep Learning (DL) model \cite{gupta2025cross}, based on the current state-of-the-art transformer architecture. More specifically, we fine-tuned the TinyViT model as an ASD classifier. Structurally, TinyViT is adapted from the hierarchical vision transformer architecture. TinyViT transformers are designed to address key constraints in Vision Transformers (ViTs) and Convolutional Neural Networks (CNNs) regarding computational efficiency, global context modelling, and inductive biases. While standard ViTs process input data as sequential patch arrays through self-attention mechanisms - enabling long-range dependencies to be captured irrespective of spatial proximity - their excessive computational requirements present deployment limitations. This challenge is mitigated through hierarchical knowledge distillation, whereby predictive capabilities from larger ViTs are transferred to compact architectures via prediction space alignment. Consequently, inference latency and memory consumption are substantially reduced while global contextual assimilation is preserved.

Simultaneously, CNN limitations stemming from fixed receptive fields and inherent spatial locality presumptions are overcome. Rigid architectural presuppositions concerning spatial relationships are avoided, permitting more intricate data representations to be learned. Progressive learning schedules are implemented, with models initially being trained at reduced resolutions before higher-dimensional fine-tuning is conducted. This enables scalable deployment across diverse hardware.

\begin{figure}[!h]
  \centering
   %{\epsfig{file = Figures/, width = 5.5cm}}
    \includegraphics[width=1\linewidth]{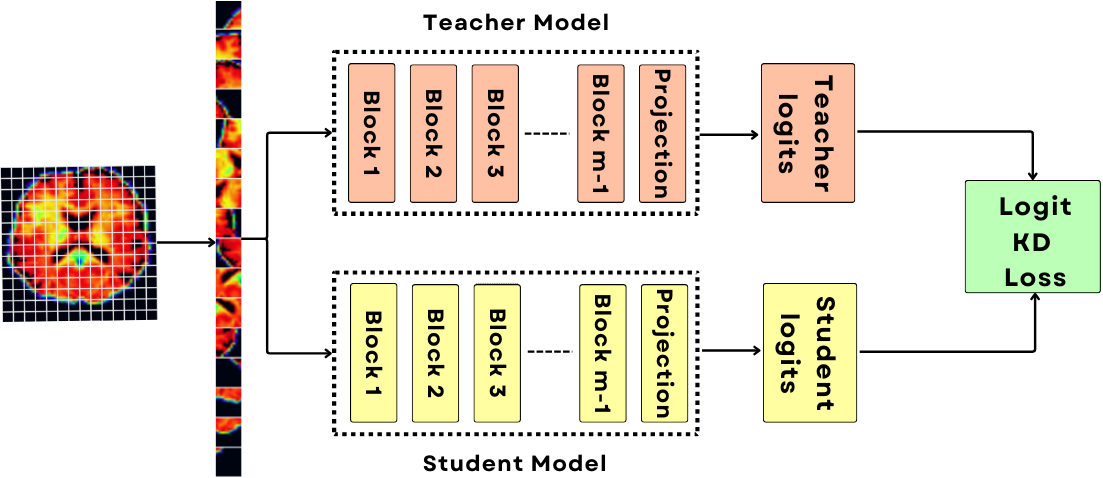}
  \caption{Overview of the implemented pre-trained knowledge distillation methodology. The upper processing pathway is dedicated to teacher logit transformation, while the lower pathway is designated for student logit computation. These top branches were fine-tuned independently. \cite{gupta2025cross}} \label{fig:archi} \centering
\end{figure}

% Knowledge transfer from the teacher model to the compact student model is facilitated via distillation within a teacher-student framework \cite{hinton2015distilling}. Teacher logits are utilized to optimize training efficiency in this process. The models employed were first trained on ImageNet21K, followed by fine-tuning on ImageNet1K. Subsequent domain adaptation was achieved by further fine-tuning on the ABIDE dataset to establish the teacher model. The implemented teacher-student architecture is illustrated in Figure (\ref{fig:archi}).

% The student model was similarly pretrained on ImageNet21K prior to distillation. Enhanced fine-tuning of the student model was driven by distillation loss ($L_{distill}$). Ultimately, the student model was optimized using a composite loss function $L_{final}$ – a regulated combination of $L_{model}$ and $L_{distill}$ (Equation \ref{eq1}). Here, $L_{model}$ denotes the student's logit loss, while $L_{distill}$ represents the Kullback-Leibler divergence \cite{chien2018source} between teacher and student logits. Through this mechanism, accelerated domain knowledge acquisition by the student model is enabled. These loss functions are formally defined as follows.

We utilized the TinyViT models pre-trained on natural image datasets to overcome the data scarcity. Knowledge transfer from the teacher model to the compact student model is facilitated via distillation within a teacher-student framework \cite{hinton2015distilling} as shown in Figure (\ref{fig:archi}). Teacher logits are utilized to optimize training efficiency in this process. The models employed were first trained on ImageNet21K, followed by fine-tuning on ImageNet1K. Subsequent domain adaptation was achieved by further fine-tuning on the ABIDE dataset to establish the teacher model. %The implemented teacher-student architecture is illustrated in Figure (\ref{fig:archi}).

Enhanced fine-tuning of the student model was driven by distillation loss ($L_{distill}$). Ultimately, the student model was optimized using a composite loss function $L_{final}$ – a regulated combination of $L_{model}$ and $L_{distill}$ (Equation \ref{eq1}). Here, $L_{model}$ denotes the student's logit loss, while $L_{distill}$ represents the Kullback-Leibler divergence \cite{chien2018source} between teacher and student logits. Through this mechanism, accelerated domain knowledge acquisition by the student model is enabled. These loss functions are formally defined as follows.

\begin{equation}\label{eq1}
L_{final} = L _{model} * \alpha + L _{distill}*(1-\alpha)
\end{equation}
%where $L_{distill}$ is defined by,
\begin{equation}
L _{distill} =  KL(P||Q)=\sum_{x}P(x)\log(\frac{P(x)}{Q(x)})
\end{equation}
and $\alpha$ \footnote{$\alpha$ = 0.5 was used through out the experiments.} was the hyper-parameter to offset the $L_{final}$ loss.

\subsection{Second module}
% The second framework module integrates three explainable AI (XAI) methodologies. Saliency maps (Attention maps), Grad-CAM, and SHAP were employed to provide further insights into decision pathways. These XAI techniques have demonstrated significant promise in medical imaging contexts—for instance, melanoma detection has been enhanced through Grad-CAM variants \cite{gamage2024melanoma}, while glaucoma diagnostics have been advanced via visualization methods. However, successful application has largely been confined to domains where clinically relevant features are visually apparent, with more complex neurological patterns presenting persistent challenges. To address this gap, our approach systematically applied these XAI methods to pinpoint critical neuroanatomical regions implicated in ASD. Identified biomarkers were then cross-referenced with established neurobiological literature, revealing reassuring convergence between computational findings and existing pathophysiological models. This validation step bridges artificial intelligence with clinical neuroscience, transforming algorithmic outputs into neurologically grounded insights. Detailed methodological descriptions follow in subsequent subsections.

Interpretability is regarded as essential within clinical AI due to the inherent "black box" nature of complex deep learning models, which frequently fail to reveal the neuroanatomical or pathophysiological basis for their diagnostic classifications. When algorithmic decisions are made without transparent reasoning, clinical validity is compromised, as established biomedical knowledge cannot be referenced to verify outputs. Consequently, trust among practitioners and patients is undermined, hindering clinical adoption. Explainable AI (XAI) methodologies address this critical gap by ensuring that diagnostic rationales are explicitly articulated, thereby transforming opaque predictions into clinically actionable intelligence. Mechanistic insights are generated through these interpretability techniques, enabling therapeutic strategies to be individualised according to an individual's condition rather than statistical correlations alone.

The second framework module integrates three explainable AI (XAI) methodologies. Saliency maps (Attention maps), Grad-CAM, and SHAP were employed to provide further insights into decision pathways. The aforementioned XAI techniques have demonstrated significant promise when compared to other XAI methods for fMRI data in medical imaging contexts. Usually single interpretability technique is employed, without comparative analysis. The results obtained from the single-explanation technique are susceptible to methodological biases and may yield incomplete or misleading rationales. 

Our approach systematically applies multiple XAI methods to pinpoint critical neuroanatomical regions implicated in ASD. These resulting neuroanatomical findings are cross-validated across distinct interpretability paradigms, providing robust analysis that is method-independent. These identified brain regions are then cross-referenced with established neurobiological literature, revealing reassuring convergence between computational findings and existing pathophysiological models. This validation step bridges artificial intelligence with clinical neuroscience, transforming algorithmic outputs into neurologically grounded insights.

\section{Experimentation}
All experiments were conducted using an NVIDIA GeForce GTX 1080 Ti GPU (12 GB RAM). To enhance dataset diversity and avoid over-fitting, strategic data augmentation techniques were employed, including centre cropping, image sharpening, controlled colour variation, and randomized contrast adjustment. Special consideration was given to demographic representation: class weighting mechanisms were carefully calibrated during training to balance ASD and neurotypical control (TC) cohorts, ensuring equitable model attention to both diagnostic categories throughout the learning process.

\subsection{First module: DL model} \label{subsec: experiment_fc}
The ViT (ViT\_B\_16) architecture was utilized to establish a baseline model by both the teacher and student models. Initial fine-tuning was performed on the ABIDE dataset for 65 epochs to establish the teacher model. Subsequently, the student model was adapted to the CMI-HBN dataset over 40 epochs using the composite loss function $L_{final}$ detailed in subsection \ref{subsec:firstcom}. Optimization parameters\footnote{Hyper-parameter optimization via Optuna \cite{optuna_2019} \label{param_opt}.} were standardized: AdamW optimizer (learning rate=3.6e-05, weight decay=1e-4) with multistep learning rate reduction (factor=0.1 every 10 epochs). 
To explore efficiency-performance tradeoffs, two compact TinyViT variants were adapted: TinyViT\_5m\_224 (5M parameters) and TinyViT\_21m\_224 (21M parameters), both processing 224×224 inputs. The smaller variant was refined on ABIDE (100 epochs) for teacher initialization, followed by student adaptation to CMI-HBN (40 epochs). Similarly, the larger variant underwent ABIDE pretraining (50 epochs) before student transfer. Distinct optimization strategies\footref{param_opt} were employed: Adam optimizer (lr=9.56e-4, wd=1e-4) with ReduceLROnPlateau scheduling (factor=0.5 after 3 epochs without validation loss improvement). The Multilayer perceptrons (MLPs) across all architectures were similarly optimized\footref{param_opt}.

Comparative benchmarking included four established CNN architectures: VGG16 \cite{simonyan2014very}, AlexNet \cite{krizhevsky2012ImageNet}, ResNet101 \cite{he2016deep}, and MobileNet \cite{howard2017mobilenets}. Identical knowledge transfer protocols (subsection \ref{subsec:firstcom}) were applied: teachers were developed through 60 epochs of ABIDE fine-tuning, while students underwent 40 epochs on CMI-HBN. Uniform hyper-parameters were maintained: Adam optimization (lr=1e-3, wd=1e-4) with decade learning rate reduction (factor=0.1 every 10 epochs).

\subsection{Second module: Explainability}
Three interpretability methods were evaluated to elucidate the "black box" model and identify neuroanatomical brain regions significant for ASD classification. Saliency mapping, Grad-CAM, and SHAP analysis were employed to quantify how the changes in the input feature set influence the prediction of the model. While saliency and Grad-CAM utilize backpropagation-based techniques, where importance scores are recursively propagated backward through network layers. SHAP leverages a game-theoretic attribution framework. The features are assigned importance scores reflecting their predictive influence: positive values indicate supportive evidence, while negative values suggest contradictory indicators, with magnitude revealing effect strength. This methodological triangulation was intentionally implemented because distinct aspects of extracted features might be emphasized by each technique.

Important feature sets were determined through these approaches and subsequently averaged. For each method, the frequency of Region of Interest (ROI) occurrence was calculated, with significant ROIs mapped to anatomical labels via Brodmann Area (BA) designations. Intersecting features across all three methodologies were prioritized, enabling isolation of key neuroanatomical regions driving ASD classification as shown in Figures (\ref{fig:saliency}, \ref{fig:grad_cam}, and \ref{fig:shap}). Finally, the identified key regions were rigorously compared against established neurobiological correlates of ASD.

\begin{table}[]
\centering
\caption{Benchmark analysis comparing our framework's performance against prior ABIDE-based methodologies.\cite{gupta2025cross}}
\label{tab:perf_compare}
%\resizebox{0.3\textwidth}{!}{%
\begin{tabular}{|c|c|}
\hline
\textbf{Studies} & \textbf{Accuracy(\%)} \\ \hline
Heinsfeld et al. \cite{heinsfeld2018identification} & 70 \\ \hline
Plitt et al. \cite{plitt2015functional} & 69.7  \\ \hline
Dvornek et al. \cite{Dvornek2017-bs} & 68.5 \\ \hline
Sherkatghanad et al. \cite{sherkatghanad2020automated} & 70.22  \\ \hline
Nielsen et al. \cite{nielsen2013multisite} & 60  \\ \hline
\textbf{Our approach} & \textbf{76.62}  \\ \hline
\end{tabular}
%}
\end{table}

\section{Results}
In this section, we will discuss the results in two folds. Firstly, the classification performance is evaluated across multiple model configurations detailed in subsection (\ref{subsec: experiment_fc}). Comprehensive results are systematically presented in Table \ref{tab:result_table}, while comparative benchmarking against prior ASD diagnostic studies is documented in Table \ref{tab:perf_compare}. Notably, the first module of our framework demonstrated superior performance relative to conventional methodologies that relied exclusively on training models de novo. These reference methods were frequently constrained by a limited dataset scale, impeding optimal performance attainment. The deep learning models in our approach were fine-tuned using cross-domain transfer learning augmented with knowledge distillation loss. This strategy leverages pretrained representations to mitigate data scarcity challenges while enhancing small-dataset generalization. As demonstrated in Table \ref{tab:result_table}, the TinyViT\_21M architecture achieved performance exceeding both ViT\_B\_16 and ViT\_B\_32 despite approximately 75\% parameter reduction. This efficiency is attributed to the hierarchical feature extraction capabilities inherent in the adapted transformer framework, which enables multi-scale representation learning not attainable through standard ViT architectures.

\begin{table*}[hbt!]
\centering
\caption{Performance comparison across transformer-based architectures. \cite{gupta2025cross}}
\label{tab:result_table}
\resizebox{\textwidth}{!}{%
\begin{tabular}{|c|c|c|c|c|c|c|c|c|}
\hline
\textbf{Models} & \textbf{Accuracy (\%)} & \textbf{Precision(\%)} & \textbf{\begin{tabular}[c]{@{}c@{}}Recall/\\ TPR(\%)\end{tabular}} & \textbf{\begin{tabular}[c]{@{}c@{}}TNR/\\ Specificity(\%)\end{tabular}} & \textbf{FPR(\%)} & \textbf{\begin{tabular}[c]{@{}c@{}}F1 \\ Score(\%)\end{tabular}} & \textbf{\begin{tabular}[c]{@{}c@{}}Model\\ Size (Million)\end{tabular}} & \textbf{\begin{tabular}[c]{@{}c@{}}Embedding \\ dim\end{tabular}} \\ \hline
VIT\_B\_16 & 72.53 & 77.35 & 63.72 & 81.33 & 18.67 & 69.88 & 86 & 768 \\ \hline
VIT\_B\_32 & 73.8 & 78.3 & 65.4 & 82.6 & 17.4 & 71.18 & 88.22 & 768 \\ \hline
TinyViT\_5m\_224 & 70.9 & 72.25 & 67.87 & 73.93 & 26.07 & 69.9 & 5 & 320 \\ \hline
\textbf{TinyViT\_21m\_224} & 76.62 & 72.23 & 86.48 & 66.75 & 33.25 & \textbf{78.72} & 21 & 576 \\ \hline
\end{tabular}%
}
\end{table*}

The results indicate that knowledge acquired from natural images is effectively adapted to fMRI data through the application of a cross-domain transfer learning approach. Enhancement of feature learning in the student model is facilitated by the guidance provided by the teacher model. These findings suggest that cross-domain transfer learning methods may offer a viable strategy for addressing challenges in data-intensive domains where sample sizes are limited. Additionally, attention-based architectures, encompassing both ViT and TinyViT across various scales, demonstrate superior performance compared to traditional CNN architectures, underscoring the advantages of transformer-based architectures. As observed in Table \ref{tab:cnn_res_table}, the performance of traditional CNN models fell below expectations. This outcome may be attributed to the limitations of CNN models in capturing global relationships within image features, which impedes the efficient transfer of specific attributes learned from the ImageNet dataset to brain imaging data.

\begin{table*}[hbt!]
\centering
\caption{Classification efficacy of convolutional neural network (CNN) variants. \cite{gupta2025cross}}
\label{tab:cnn_res_table}
\resizebox{\textwidth}{!}{%
\begin{tabular}{|c|c|c|c|c|c|c|}
\hline
\textbf{Models} & \textbf{Accuracy(\%)} & \textbf{Precision(\%)} & \textbf{\begin{tabular}[c]{@{}c@{}}Recall/\\ TPR(\%)\end{tabular}} & \textbf{\begin{tabular}[c]{@{}c@{}}TNR/\\ Specificity(\%)\end{tabular}} & \textbf{FPR(\%)} & \textbf{\begin{tabular}[c]{@{}c@{}}F1 \\ Score(\%)\end{tabular}} \\ \hline
VGG16 & 64.3 & 67.2 & 59.3 & 38.5 & 61.05 & 58.12 \\ \hline
Alexnet & 60.6 & 62.8 & 57.2 & 40.2 & 58.6 & 59.86 \\ \hline
Resnet101 & 67.3 & 70.2 & 60.6 & 64.4 & 39.8 & 65.06 \\ \hline
MobileNet & 66.8 & 69.4 & 59.2 & 60.3 & 42.6 & 63.89 \\ \hline
\end{tabular}%
}
\end{table*}

In a contrastive analysis conducted between the TinyViT\_5M model and its counterparts, ViT\_B\_16 and ViT\_B\_32. The TinyViT\_5M model, characterized by a modest parameter count of 5 million, was found to exhibit performance levels akin to those observed in the ViT\_B\_16 model, despite the latter being equipped with a substantially higher parameter count of 86 million. This equivalence in performance is attributed to the efficient utilization of parameters within the TinyViT\_5M architecture, which has been optimized to extract critical features effectively despite its reduced scale. Furthermore, it was revealed that no notable enhancement in performance was achieved when the ViT\_B\_32 model, possessing an even greater number of parameters than ViT\_B\_16, was employed. This lack of improvement is likely influenced by the constrained size of the datasets utilized in the study. With limited data available, the essential features appear to have been largely captured by the models, leaving minimal opportunity for additional insights to be gained by the larger ViT\_B\_32 configuration. 

\begin{figure*}[!h]
  \centering
   %{\epsfig{file = Figures/, width = 5.5cm}}
    \includegraphics[width=\linewidth]{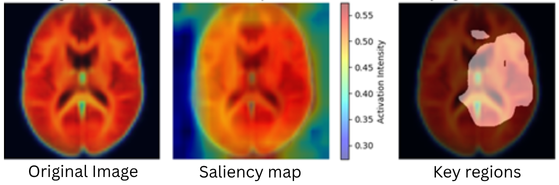}
  \caption{The original rs-fMRI scan (left), generated saliency map (middle), and significant regions of interest (ROIs) highlighted in the right panel.} \label{fig:saliency} \centering.
\end{figure*}

\begin{figure*}[!h]
  \centering
   %{\epsfig{file = Figures/, width = 5.5cm}}
    \includegraphics[width=\linewidth]{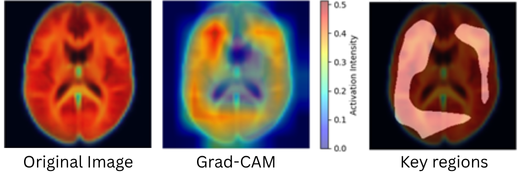}
  \caption{Visualization of Grad-CAM methodology, original scan(left), Gradient-weighted Class Activation Mapping output (middle), significant regions of interest (ROIs) are highlighted in the right.} \label{fig:grad_cam} \centering.
\end{figure*}

\begin{figure*}[!h]
  \centering
   %{\epsfig{file = Figures/, width = 5.5cm}}
    \includegraphics[width=\linewidth]{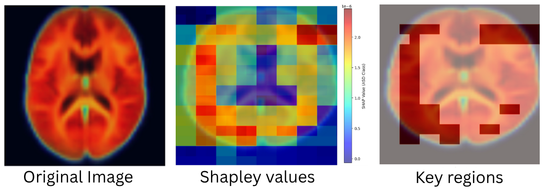}
  \caption{SHAP output: original fMRI scan (left), shapley values overlaid onto the original scan, highlighting regions and their associated importance score, critical brain ROIs identified as most significant by the method.} \label{fig:shap} \centering.
\end{figure*}

Secondly, three explainable AI (XAI) methodologies —namely, saliency mapping, Grad-CAM, and SHAP analysis —were strategically deployed to provide insights into the model's diagnostic pathways and pinpoint neurofunctionally critical regions. Discriminative features were identified through saliency mapping in Figure (\ref{fig:saliency}), with clinically significant regions catalogued in Table \ref{tab:saliency_res}. Similarly, Grad-CAM outputs were visualized in Figure (\ref{fig:grad_cam}) while neurobiological substrates were systematically documented in Table \ref{tab:grad_cam_res}. SHAP interpretation further revealed decision-informative regions through visual analytics as demonstrated in Figure (\ref{fig:shap}), and findings are tabulated in Table \ref{tab:shap_res}. Crucially, we identified consensus Broadmann Areas (BA) emerging across Tables \ref{tab:saliency_res}-\ref{tab:shap_res}, which are discriminative for the classification of ASD. This methodological triangulation yielded consistent neuroanatomical regions that clinicians can confidently associate with ASD pathology.

The consensus was observed across all three interpretability methods: the calcarine sulcus and cuneus (BA 17) were consistently identified as neurofunctionally critical. This primary visual cortex region serves as our visual gateway—where retinal signals are transformed into the edges, colours, and contours that construct our perceived world. Clinicians might recognize how disruptions here could fragment a patient's sensory experience. Further alignment emerged between saliency maps and Grad-CAM at the insula (BA 13 \& 16). This region manages the sensory inputs, emotional states, and decision-making. When people with ASD struggle with social interactions, we can relate how this region integrates bodily sensations with emotional meaning. Both saliency maps and SHAP similarly confirmed parietal lobe engagement (BA 5). It is responsible for synthesizing touch and vision to navigate physical space. Finally, Grad-CAM and SHAP jointly highlighted the middle/inferior temporal gyri (BA 21 \& 20). These linguistic and memory hubs weave words, meanings, and visual perceptions into a coherent understanding. These regions are responsible for why autistic individuals often experience language-processing challenges. 

These overlapping regions transform algorithmic outputs into clinically actionable intelligence, bridging artificial intelligence with neuropsychiatric expertise through transparent decision trails.

% \begin{table}[]
% \centering
% \caption{The top brain regions identified through the saliency map, along with the key regions and their associated Brodmann areas.}
% \label{tab:saliency_res}
% \begin{tabular}{ccc}
% \hline
% Identified top regions & Key regions & \begin{tabular}[c]{@{}c@{}}Corresponding \\ Brodmann's Area\end{tabular} \\ \hline
% Insula & Insula & BA 13 \& 16 \\ \hline
% Claustrum & Claustrum & - \\ \hline
% Parietal lobe & Parietal lobe & BA 5 \\ \hline
% Thalamus & Thalamus & - \\ \hline
% Temporal lobe & Temporal lobe & BA 15 \\ \hline
% \begin{tabular}[c]{@{}c@{}}Calcarine sulcus\\ (Occipital lobe)\end{tabular} & \begin{tabular}[c]{@{}c@{}}Calcarine sulcus\\ (Occipital lobe)\end{tabular} & BA 17 \\ \hline
% Cuneus & Cuneus & BA 17 \\ \hline
% \end{tabular}
% \end{table}

\begin{table}[]
\centering
\caption{The top brain regions identified through the saliency map, along with the key regions and their associated Brodmann areas.}
\label{tab:saliency_res}
\begin{tabular}{|c|c|c|}
\hline
Identified top regions & Key regions & \begin{tabular}[c]{@{}c@{}}Corresponding \\ Brodmann's Area\end{tabular} \\ \hline
Insula & Insula & BA 13 \& 16 \\ \hline
Claustrum & Claustrum & - \\ \hline
Parietal lobe & Parietal lobe & BA 5 \\ \hline
Thalamus & Thalamus & - \\ \hline
Temporal lobe & Temporal lobe & BA 15 \\ \hline
\begin{tabular}[c]{@{}c@{}}Calcarine sulcus\\ (Occipital lobe)\end{tabular} & \begin{tabular}[c]{@{}c@{}}Calcarine sulcus\\ (Occipital lobe)\end{tabular} & BA 17 \\ \hline
Cuneus & Cuneus & BA 17 \\ \hline
\end{tabular}
\end{table}

% \begin{table}[]
% \centering
% \caption{Neuroanatomically regions identified using the Gradient-weighted Class Activation Mapping (Grad-CAM) method, key regions with corresponding Brodmann areas.}
% \label{tab:grad_cam_res}
% \begin{tabular}{ccc}
% \hline
% Identified top regions & Key regions & \begin{tabular}[c]{@{}c@{}}Corresponding \\ Brodmann's Area\end{tabular} \\ \hline
% Middle frontal gyrus & Middle frontal gyrus & - \\ \hline
% Temporal gyrus & \begin{tabular}[c]{@{}c@{}}Middle temporal gyrus \&\\  Inferior temporal gyrus\end{tabular} & BA 21 \& BA 20 \\ \hline
% \begin{tabular}[c]{@{}c@{}}Calcarine sulcus\\ (Occipital lobe)\end{tabular} & \begin{tabular}[c]{@{}c@{}}Calcarine sulcus\\ (Occipital lobe)\end{tabular} & BA 17 \\ \hline
% Cuneus & Cuneus & BA 17 \\ \hline
% Insula & Insula & BA 13 \& BA 16 \\ \hline
% \end{tabular}
% \end{table}

\begin{table}[]
\centering
\caption{Neuroanatomically regions identified using the Gradient-weighted Class Activation Mapping (Grad-CAM) method, key regions with corresponding Brodmann areas.}
\label{tab:grad_cam_res}
\begin{tabular}{|c|c|c|}
\hline
Identified top regions & Key regions & \begin{tabular}[c]{@{}c@{}}Corresponding \\ Brodmann's Area\end{tabular} \\ \hline
Mid. frontal gyrus & Mid. frontal gyrus & - \\ \hline
Temporal gyrus & \begin{tabular}[c]{@{}c@{}}Mid. temporal gyrus \&\\ Inf. temporal gyrus\end{tabular} & BA 21 \& BA 20 \\ \hline
\begin{tabular}[c]{@{}c@{}}Calcarine sulcus\\ (Occipital lobe)\end{tabular} & \begin{tabular}[c]{@{}c@{}}Calcarine sulcus\\ (Occipital lobe)\end{tabular} & BA 17 \\ \hline
Cuneus & Cuneus & BA 17 \\ \hline
Insula & Insula & BA 13 \& BA 16 \\ \hline
\end{tabular}
\end{table}

% \begin{table}[]
% \centering
% \caption{The significant regions isolated through SAHP analysis are presented, with key regions mapped to their corresponding Brodmann areas.}
% \label{tab:shap_res}
% \begin{tabular}{ccc}
% \hline
% Identified top regions & Key regions & \begin{tabular}[c]{@{}c@{}}Corresponding \\ Brodmann's Area\end{tabular} \\ \hline
% Sup. temporal gyrus & Sup. temporal gyrus & BA 22 \\ \hline
% \begin{tabular}[c]{@{}c@{}}Calcarine sulcus\\ (Occipital lobe)\end{tabular} & \begin{tabular}[c]{@{}c@{}}Calcarine sulcus\\ (Occipital lobe)\end{tabular} & BA 17 \\ \hline
% Cuneus & Cuneus & BA 17 \\ \hline
% Temporal gyrus & \begin{tabular}[c]{@{}c@{}}Mid. temporal gyrus \&\\ Inf. temporal gyrus\end{tabular} & BA 21 \& BA 20 \\ \hline
% Parietal lobe & Parietal lobe & BA 5 \\ \hline
% \end{tabular}
% \end{table}

\begin{table}[]
\centering
\caption{The significant regions isolated through SAHP analysis are presented, with key regions mapped to their corresponding Brodmann areas.}
\label{tab:shap_res}
\begin{tabular}{|c|c|c|}
\hline
Identified top regions & Key regions & \begin{tabular}[c]{@{}c@{}}Corresponding \\ Brodmann's Area\end{tabular} \\ \hline
Sup. temporal gyrus & Sup. temporal gyrus & BA 22 \\ \hline
\begin{tabular}[c]{@{}c@{}}Calcarine sulcus\\ (Occipital lobe)\end{tabular} & \begin{tabular}[c]{@{}c@{}}Calcarine sulcus\\ (Occipital lobe)\end{tabular} & BA 17 \\ \hline
Cuneus & Cuneus & BA 17 \\ \hline
Temporal gyrus & \begin{tabular}[c]{@{}c@{}}Mid. temporal gyrus \&\\ Inf. temporal gyrus\end{tabular} & BA 21 \& BA 20 \\ \hline
Parietal lobe & Parietal lobe & BA 5 \\ \hline
\end{tabular}
\end{table}

\section{Discussion}
The first module in our proposed framework employs a cross-domain transfer learning methodology. Within this module, pre-trained TinyViT and ViT models underwent fine-tuning utilizing a teacher-student paradigm combined with knowledge distillation techniques. Conversely, the comparative methods detailed in Table \ref{tab:perf_compare} relied on conventional machine learning strategies, involving model training initiated from the ground up. These comparative techniques frequently encounter limitations stemming from the dataset's constrained scale and inherent difficulties in capturing essential feature representations effectively, often yielding insufficiently robust outcomes. To mitigate these constraints, adaptation of pre-trained TinyViT models to the target dataset was implemented through fine-tuning.

The utilization of pre-trained TinyViT architectures offers several distinct advantages. Primarily, the facilitation of knowledge transfer from extensive natural image datasets to the specialized domain of brain imaging is enabled by cross-domain transfer learning and knowledge distillation. Consequently, enhanced feature acquisition capabilities were consistently observed. Secondly, the hierarchical transformer-based structure intrinsic to TinyViT facilitates the processing of images as sequential patch arrays via window-based attention mechanisms. This characteristic permits the consideration of interdependencies among patches regardless of their spatial separation, thereby improving the model's capacity to assimilate long-range contextual information and global dependencies. Furthermore, a reduced structural preconception is conferred upon TinyViT models when contrasted with convolutional neural networks (CNNs). Unlike CNNs, which are constrained by presumptions of locality in spatial configurations, TinyViT architectures are not bound by such presuppositions, permitting the learning of more intricate and abstract data representations. Additionally, the substantially reduced parameter count relative to alternative hierarchical transformers and conventional Vision Transformers (ViTs) renders TinyViT models particularly advantageous for contexts involving limited datasets, ensuring both computational economy and adaptability. This combination of collective characteristics positions TinyViT as optimally aligned with the methodological requirements of the proposed approach.

The consensus of visual processing regions across all interpretability methods provides convincing evidence for its major role in ASD diagnosis. The convergence observed on primary visual cortical regions—specifically the calcarine sulcus and cuneus, corresponding to Broadmann area 17—may reflect a core characteristic transcending methodological variations in feature interpretation. The prominence of BA 17 (primary visual cortex) within these findings carries particular weight, given its independent validation over diverse research areas. The importance of BA (17) in autism has been highlighted by genetic investigations utilizing distinct models and datasets \cite{gandal2022broad}, whereas neurophysiological findings have demonstrated that deficiencies in motion perception \cite{robertson2014global} and atypical oscillatory activity (e.g., gamma oscillations) \cite{orekhova2023gamma} are related to this region in ASD. The identification of these regions by the presented model thus corroborates a growing recognition that the occurrence of these elemental disparities in visual perception contributes as a critical factor in ASD pathophysiology.

Similarly, identification of the cuneus across the three interpretability methods corresponds with recent findings, where diminished connectivity between brainstem and cuneus regions has been observed in autism cohorts relative to their typically developing co-twins \cite{cheng2015autism}. Such alterations in brain connectivity within lower-level visual pathways are understood to impact both foundational perceptual capabilities and the processing of socially relevant information, suggesting a neural pathway through which early sensory processing may influence higher-order social characteristics. Further corroboration is provided by resting-state functional magnetic resonance imaging (rs-fMRI) investigations and eye-tracking studies, which have equivocally associated cuneus activity patterns with social processing differences observed in autism \cite{xiao2023atypical}. Additional corroboration is afforded by the concurrent identification of the middle and inferior temporal gyrus BA (21 \& 20) through both Grad-CAM and SHAP methodologies. This validation derives from these regions' well-documented involvement in linguistic functions, semantic memory formation, and visual interpretation, alongside the characteristic communicative challenges observed in ASD \cite{monk2009abnormalities}.

Concurrent validation of parietal lobe involvement BA(5) was achieved through both saliency map analysis and SHAP methodologies. This region is implicated in the regulation of sensory perception and spatial reasoning. Analysis conducted in the study \cite{travers2015motor} indicated reduced efficacy in motor sequence acquisition among individuals with ASD. Neuroimaging data revealed diminished activation within BA (5) during learning tasks when ASD cohorts were compared with neurotypical participants. Furthermore, increased severity of repetitive behavioural patterns and restricted interests among ASD participants was correlated with more pronounced activation reductions in these parietal regions.

Further concordance was observed between saliency maps and Grad-CAM outputs within the insular cortex BA (13 \& 16). This region is implicated in the regulation of sensory integration, emotive states, and decision formation. Difficulties encountered during social engagement by individuals with ASD can be conceptually linked to how this area synthesizes interoceptive signals with emotional significance \cite{caria2015anterior}. Findings from the study \cite{ebisch2011altered} indicated diminished functional coupling in ASD cohorts relative to typically developing (TD) groups. This reduced connectivity involved both anterior and posterior insular subdivisions and specific neural structures dedicated to affective and sensory processing.
The alignment between language-associated and visual processing areas implies that the diagnosis of ASD is mediated by a distributed neural architecture encompassing both primary sensory pathways and advanced cognitive systems.

The adoption of the proposed methodology within clinical environments would reduce reliance on conventional diagnostic instruments such as ADOS scoring systems, while diminishing the necessity for repeated patient evaluations to ensure diagnostic reliability. Clinicians would be provided with actionable resources through computer-aided diagnostic (CAD) systems developed from this framework, enabling more precise and expedient assessments while facilitating evidence-based clinical decisions. Furthermore, diagnostic workflows could be streamlined through such integration, allowing for a greater emphasis on individualized intervention strategies and a deeper exploration of autism’s neurologically grounded mechanisms.

\section{Conclusion}
This study has introduced a novel dual-module framework designed to advance both the accuracy and interpretability of autism spectrum disorder (ASD) diagnosis. The first module leveraged cross-domain transfer learning and knowledge distillation to fine-tune compact, hierarchical vision transformers (Tiny ViT) for fMRI-based classification. This approach effectively addressed data scarcity challenges, achieving superior performance (76.62\% accuracy) compared to conventional CNNs and larger transformer variants. The computational efficiency and parameter economy of TinyViT—coupled with its capacity to model long-range dependencies and abstract feature representations—demonstrated significant advantages for neuroimaging applications with limited datasets.

The second module integrated three complementary explainable AI (XAI) techniques—saliency mapping, Grad-CAM, and SHAP analysis—to elucidate the model’s diagnostic pathways and identify neurofunctionally critical brain regions. A robust consensus emerged across methods, highlighting the central involvement of primary visual processing regions (calcarine sulcus and cuneus, BA 17), the insula (BA 13 \& 16), parietal lobe (BA 5), and middle/inferior temporal gyri (BA 21 \& 20). This convergence not only validated the model’s alignment with established neurobiology but also revealed a distributed neural architecture underpinning ASD pathophysiology. Critically, the prominence of early visual processing regions (BA 17) corroborates growing evidence of sensory integration deficits in ASD, while connectivity aberrations in the insula and parietal lobe provide mechanistic insights into social and repetitive behavioural symptoms.

The triangulation of XAI findings with independent genetic, neurophysiological, and connectivity studies underscores the clinical validity of this framework. By transforming opaque model decisions into neurologically grounded explanations, our approach bridges artificial intelligence with clinical neuroscience, offering a transparent, interpretable tool for practitioners. Future work will focus on validating these biomarkers across diverse cohorts and integrating multimodal data to further refine diagnostic precision. Ultimately, this framework advances the development of clinically actionable CAD systems, fostering earlier intervention and personalized management strategies for ASD.

%
% the environments 'definition', 'lemma', 'proposition', 'corollary',
% 'remark', and 'example' are defined in the LLNCS documentclass as well.
%

% For citations of references, we prefer the use of square brackets
% and consecutive numbers. Citations using labels or the author/year
% convention are also acceptable. The following bibliography provides
% a sample reference list with entries for journal
% articles \cite{ref_article1}, an LNCS chapter \cite{ref_lncs1}, a
% book~\cite{ref_book1}, proceedings without editors~\cite{ref_proc1},
% and a homepage~\cite{ref_url1}. Multiple citations are grouped
% \cite{ref_article1,ref_lncs1,ref_book1},
% \cite{ref_article1,ref_book1,ref_proc1,ref_url1}.

\begin{credits}
\subsubsection{\ackname} %A bold run-in heading in small font size at the end of the %\section{\uppercase{Acknowledgements}}

We want to thank EPSRC DTP HMT for funding this project. Also, this manuscript was prepared using a limited-access dataset obtained from the Child Mind Institute Biobank, HBN dataset. This manuscript reflects the views of the authors and does not necessarily reflect the opinions or views of the Child Mind Institute.
\end{credits}
\bibliographystyle{splncs04}
{\small
\bibliography{main}}%

\end{document}